\documentclass[runningheads]{llncs}
\usepackage[T1]{fontenc}
\usepackage{graphicx}
\usepackage{color}
\usepackage{times}
\usepackage{latexsym}
\usepackage[T1]{fontenc}
\usepackage[utf8]{inputenc}
\usepackage{microtype}
\usepackage{multirow}
\usepackage{graphicx}
\usepackage{lipsum}
\usepackage{blindtext}
\usepackage{makecell}
\usepackage{amsmath}
\usepackage{bbm}
\usepackage{bm}
\usepackage{CJKutf8} 
\usepackage{appendix}
\usepackage{etoolbox}
\usepackage{tikz}
\def\checkmark{\tikz\fill[scale=0.3](0,.4) -- (.25,0) -- (0.8,.7) -- (.25,.2) -- cycle;} 

\usepackage[numbers,square,sort&compress]{natbib}

\begin{document}
\title{D2CSE: Difference-aware Deep continuous prompts for Contrastive Sentence Embeddings}
\titlerunning{Difference-aware Deep continuous prompts for Contrastive Sentence Embeddings}
%
%
%
\author{Hyunjae Lee}
%
\institute{Samsung SDS, Korea \\
\email{h8.lee@samsung.com}
}
\maketitle              
\begin{abstract}
This paper describes Difference-aware Deep continuous prompt for Contrastive Sentence Embeddings (D2CSE) that learns sentence embeddings. Compared to state-of-the-art approaches, D2CSE computes sentence vectors that are exceptional to distinguish a subtle difference in similar sentences by employing a simple neural architecture for continuous prompts. Unlike existing architectures that require multiple pretrained language models (PLMs) to process a pair of the original and corrupted (subtly modified) sentences, D2CSE avoids cumbersome fine-tuning of multiple PLMs by only optimizing continuous prompts by performing multiple tasks---i.e., contrastive learning and conditional replaced token detection all done in a self-guided manner. D2CSE overloads a single PLM on continuous prompts and greatly saves memory consumption as a result. The number of training parameters in D2CSE is reduced to about 1\% of existing approaches while substantially improving the quality of sentence embeddings. We evaluate D2CSE on seven Semantic Textual Similarity (STS) benchmarks, using three different metrics, namely, Spearman's rank correlation, recall@K for a retrieval task, and the anisotropy of an embedding space measured in alignment and uniformity. Our empirical results suggest that shallow (not too meticulously devised) continuous prompts can be honed effectively for multiple NLP tasks and lead to improvements upon existing state-of-the-art approaches.


\keywords{Sentence Embeddings  \and Sentence Retrieval  \and Contrastive Learning \and Continuous Prompts.}
\end{abstract}

\section{Introduction}
In natural language processing (NLP), sentence representation learning maps text of a natural language sentence onto a vector in a semantic embedding space. A good representation captures informative, linguistic features of a sentence and can be used to distinguish subtle differences of similar text. Sentence embedding is a vital component in contemporary NLP~\cite{conneau,mishra2019survey,li2022brief} and has shown the effectiveness in various tasks such as sentence classification, generation, and semantic search. Especially the quality of sentence embeddings plays a crucial role in text retrieval tasks. Recently, contrastive learning approaches are found beneficial to learn sentence representation. Combined with data augmentation methods in particular, one can conveniently learn high-performance sentence embeddings in an unsupervised manner. \citet{simcse} propose SimCSE that uses dropout to generate randomly perturbed sentence vectors for the same textual input. Numerous approaches~\cite{esimcse,CMLM,sncse,diffcse,dcpcse,zhang2022mcse} have followed SimCSE successfully.

Several approaches~\cite{equivariant,diffcse} point out the problem of invariant contrastive learning like SimCSE that the representation is only tailored to be insensitive to any perturbations of embeddings.
Especially, DiffCSE directly inspired by equivariant contrastive learning~\cite{equivariant} rather encourages representation to be sensitive to semantically meaningful transformation by employing the discriminator of ELECTRA~\cite{clark2019electra} with a novel conditional replaced token detection (CRTD) task.
In their framework, a discriminator distinguishes original tokens from plausible replacements in a input sentence by conditioning on the sentence representation produced by a sentence encoder.
An encoder thereby learns to produce rich enough embeddings for a discriminator to discern slight nuances among similar sentences.
However, these framework inevitably requires an additional pretrained language model (PLM) like BERT~\cite{devlin2019bert} as a discriminator, resulting in doubled training parameters and resources. In addition, a discriminator is not involved in producing sentence embeddings, so it becomes dispensable in inference phase.

We propose D2CSE or \underline{D}ifference-aware \underline{D}eep continuous prompts for \underline{C}ontrastive \underline{S}entence \underline{E}mbeddings, a contrastive learning framework for sentence embeddings that sets up continuous prompts to complement equivariant contrastive learning.
Continuous prompts consist of continuous vectors whose dimension is the same as the hidden layer of a PLM and are prepended to each layer's hidden representations. 
Then, we only fine-tune continuous prompts along with a dense layer on top of encoder and discriminator as a classifier for contrastive learning and CRTD task while freezing all PLM's weights.
Therefore, D2CSE reduces the number of training parameters drastically to around 1\% while learning the representation aware of both semantic similarity and a subtle difference in sentences by the combined contrastive and replaced token detection objective.

We have evaluated D2CSE empirically, using 7 semantic textual similarity (STS) datasets~\cite{conneau-kiela-2018-senteval} under both unsupervised and supervised settings with three different metrics, Spearman’s rank correlation, recall@K of the retrieval task, and the anisotropy in embedding space measured by the alignment and uniformity. 
By the STS performance and qualitative analysis, we demonstrate D2CSE successfully benefits of both continuous prompt architecture and equivariant contrastive learning methods of DiffCSE.
Our additional finding is that \texttt{[CLS]} prompting which replaces the static \texttt{[CLS]} token embedding of PLM with trainable continuous prompts promotes a more uniform distribution of embeddings so as to alleviate an anisotropic issue degrading the quality of sentence embeddings.

\section{Related Work}
\paragraph{\textbf{Contrastive learning for text embeddings.}} It is intuitive to learn good semantic representation by placing positive training examples closer, and negatives apart. Pairing positive and negative text examples to build a training corpus, however, remains a demanding challenge in NLP~\cite{cline,simcse}. Several approaches~\cite{declutr,iter2020pretraining,CMLM} leverage adjacent text segments like words, sentences, and paragraphs within a given context to automate the pairing. Other approaches resort to data augmentation methods including back-translation~\cite{cert}, deformation of text~\cite{consert}, dropout-based perturbation~\cite{simcse}, word repetition~\cite{esimcse}, and negation of the original sentences~\cite{sncse}. 

In the mean time, one can integrate an intrinsic task from existing pre-training techniques such as replaced token detection~\cite{diffcse} and masked language modeling~\cite{CMLM}, into sentence representation learning. They revamp the learning objective for sentence representation from simply contrasting sentences to generating an informative, useful conditional input as a hint for the aforementioned tasks.
\paragraph{\textbf{Continuous prompt learning.}}
Prompt learning is a promising paradigm in natural language processing, which gradually replaces a dominant role of \textit{pre-train and fine-tuning} paradigm~\cite{liu2021pre}.
While fine-tuning approach to perform downstream tasks predicts a output $\bm{y}$ from $P(\bm{y}|\bm{x})$ where $\bm{x}$ is an input and trains $P$, discrete prompt learning finds a best template $\bm{x;x'}$ or $\bm{x''}$ as a new textual input rather then training $P$. 
Inspired by this, PromptBERT~\cite{promptbert} and SNCSE~\cite{sncse} use discrete prompts to improve sentence representation. 

In contrast to discrete prompt learning, continuous prompt learning happens in the embedding space for data input~\cite{liu2021pre}.
\citet{gptunder,ptuning,lester2021power,qin2021learning,cptuning,prefix} show that continuous prompt learning, if properly set up, is comparable or even superior to supervised, task-specific fine-tuning of a downstream NLP task. 
DCPCSE~\cite{dcpcse} and Prefix-tuning~\cite{prefix} add continuous vectors to each transformer encoder layer as a prompt to an input sequence such that the prompt directly intervenes all encoding layers.
Especially, DCPCSE has hinted on a continuous prompt learning technique integrated into a contrastive learning framework. While fine-tuning only 0.1\% of the PLM parameters, DCPCSE greatly outperforms SimCSE.
However, we observe that DCPCSE fails to discern slight nuances among similar sentences from our empirical study (as described in \ref{sec:exp}) unlike DiffCSE, D2CSE.
\section{D2CSE Framework}
\begin{figure*}[!t]
\centering
\includegraphics[width=1.0\textwidth]{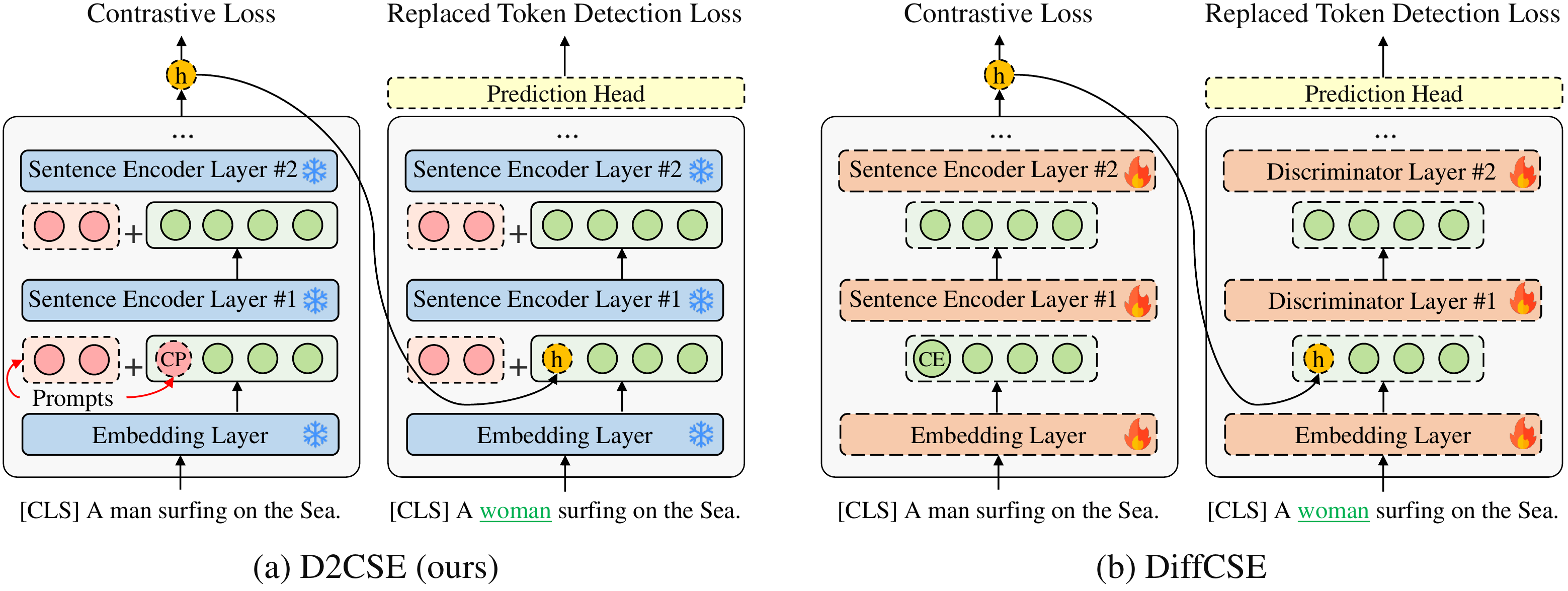}
\caption{Architectural comparison of our D2CSE and DiffCSE. 
The dashed components represent trainable units. `CP' and `CE' refer to \texttt{[CLS]} prompt and \texttt{[CLS]} token embedding, respectively.}
\label{fig:approaches1}
\end{figure*}

Figure~\ref{fig:approaches1} presents an architectural comparison of D2CSE to the baseline DiffCSE. We apply a continuous prompt encoder to DiffCSE's sentence encoder and discriminator, keeping their weight parameters fixed. 
In D2CSE, a frozen sentence encoder with a trainable continuous prompts performs replaced token detection task conditioning on the sentence vector $\mathbf{h}$ produced by itself. Therefore, continuous prompts are optimized by multiple tasks in self-guided manner. 
The prompt encoder produces a set of vectors $\mathbf{v} \in \mathbbm{R}^{a \times b \times c}$, where $a$, $b$ and $c$ denotes the number of encoder layers, a length of prompts, and the dimension of encoder layer, respectively.
These vectors are concatenated to each layer's hidden vector and optimized independently as in DCPCSE~\cite{dcpcse}. In addition, we replace the sentence encoder's static \texttt{[CLS]} token embedding vector with single tunable prompt vector, i.e., $e(\mathrm{\texttt{[CLS]}}) \leftarrow p(\mathrm{\texttt{[CLS]}}) \in \mathbbm{R}^d$, where $p$, $e$ and $d$ refer to prompt encoder, embedding matrix and its dimension, respectively. And we dub this ``$\texttt{[CLS]}$ prompt''.
We provide a detailed ablation study of different architectures in Experiments~\ref{sec:ablationstudy}.
 

\paragraph{\textbf{Contrastive learning (CL).}} The $i$th data example ${X_{i}}$ consists of $(x_{i}, x_{i}^+, x_{i}^-)$ where $x_{i}^+$ and ${x_{i}^-}$ are a positive and a negative pair for the anchor sentence ${x_{i}}$. We follow the SimCSE unsupervised setting that makes ${x_{i}^+ = x_{i}}$ and ${x_{i}^-=}$ $\emptyset$. For supervised setting, ${x_{i}^+}$ and ${x_{i}^-}$ are entailment and contradiction hypotheses of the premise ${x_{i}}$. The sentence encoder $f$ uses BERT~\cite{devlin2019bert} or RoBERTa~\cite{liu2019roberta} as a PLM. We then acquire the representation $\mathbf{h} = f_{\theta, \phi}(\mathrm{concat}(p(\text{\texttt{[CLS]}}),e(x)))$ for an input ${x}$, where only prompt encoder's parameter ${\phi}$ and $p$ are tunable while PLM's parameter $\theta$ and embedding matrix ${e}$ remain unchanged during training. The CL objective $\mathcal{L}^i_{CL}$ is:
\begin{equation}
\label{eq:cl}
{-\log\frac{e^{\mathrm{sim(h_i,h_i^+)}/\tau}}{\sum_{j=1}^{N}(e^{\mathrm{sim(h_i,h_j^+)}/\tau}+\mathbbm{1}_{[x_{\cdot}^- \neq \emptyset]}e^{\mathrm{sim(h_i,h_j^-)}/\tau})}}
\end{equation} where N is the batch size, $\tau$ is a temperature value and $\mathrm{sim(\cdot,\cdot)}$ is the cosine similarity function.

\paragraph{\textbf{Conditional replaced token detection (CRTD) learning.}}
An RTD task originally aims to train a discriminator to discriminate real tokens from plausible replacements. On the contrary, conditional RTD aims to train a sentence encoder to offer $\mathbf{h}$ as a hint for a discriminator so as to encourage $\mathbf{h}$ more informative.
Given a sentence $x$ of token length $L$, $x$ is $\{t_{1},t_{2}, ...,t_{L}\}$. Then, randomly selected $M$ tokens are replaced with plausible replacements by the generator,\footnotemark\footnotetext{We omit such generator in Figure~\ref{fig:approaches1}, because sentence generation can be done before training. We use a distilled version of BERT~\cite{sanh2019distilbert} as a generator.} which is another PLM.
Let $x_g$ is generated sentence and ${t'}$ is a token in $x_g$, then the CRTD training objective $\mathcal{L}_{\mathrm{CRTD}}(x, x_g, \mathbf{h}, f)$ \footnotemark\footnotetext{We omit the prediction head for the simplicity.} is: 
\begin{equation}
\sum_{k={1}}^{L} \left( -\mathbbm{1}\left( t'_{k}=t_{k} \right) \log f(x_g,\mathbf{h}, k)-\mathbbm{1}\left( t'_{k}\neq t_{k} \right) \log(1-f(x_g,\mathbf{h}, k)) \right)
\end{equation} where the fixed sentence encoder ${f}$ is replaced with a trainable discriminator ${D}$ in original CRTD. 
That is, ${f}$ in the above equation is represented as ${D}$, i.e., $\mathcal{L}_{\mathrm{CRTD}}(x, x_g, \mathbf{h}, D)$.
And, because we train CRTD under both supervised and unsupervised setting, more precise objective under supervised setting is $\sum_{i}\mathcal{L}_{\mathrm{CRTD}}(x^i, {x_g}^i, \mathbf{h}^i, f)$ where $i \in \{Null,+,-\}$.

\begin{table*}[ht!]
\caption{The performance comparison on STS test datasets in unsupervised setting (Spearman's rank correlation $\times$ 100. $\dagger$: results from~\cite{simcse}; $\ddagger$: results from corresponding papers.)}
\centering
\resizebox{\textwidth}{!}{
\begin{tabular}{lcccccccc}
\hline
Model                     & STS12          & STS13          & STS14          & STS15          & STS16          & STS-B          & SICK-R         & Avg.            \\ 
\hline
GloVe embeddings~(avg.)~$\dagger$   & 55.14          & 70.66          & 59.73          & 68.25          & 63.66          & 58.02          & 53.76          & 61.32           \\
BERT$_{base}$-flow~$\dagger$   & 58.40          & 67.10          & 60.85          & 75.16          & 71.22          & 68.66          & 64.47          & 66.55           \\
ConSERT-BERT$_{base}$~$\ddagger$& 64.64          & 78.49          & 69.07          & 79.72          & 75.95          & 73.97          & 67.31          & 72.74           \\
SimCSE-BERT$_{base}$~$\dagger$        & 68.40          & 82.41          & 74.38          & 80.91          & 78.56          & 76.85          & \textbf{72.23}          & 76.25           \\
ESimCSE-BERT$_{base}$~$\ddagger$ & 73.40          & 83.27          & 77.25          & 82.66          & 78.81          & 80.17          & 72.30          & 78.27           \\
DCPCSE-BERT$_{base}$~$\ddagger$       & 73.03          & \textbf{85.18} & 76.70          & 84.19 & 79.69          & 80.62          & 70.00          & 78.49           \\
DiffCSE-BERT$_{base}$~$\ddagger$    & 72.28          & 84.43          & 76.47          & 83.90          & 80.54          & 80.59          & 71.23          & 78.49           \\
PromptBERT$_{base}$~$\ddagger$~& 71.56          & 84.58          & 76.98          & \textbf{84.47}          & 80.60          & 81.60          & 69.87          & 78.54           \\
D2CSE-BERT$_{base}$         & \textbf{73.42} & 84.34          & \textbf{77.43} & 84.18          & \textbf{81.16} & \textbf{81.97} & 70.61          & \textbf{79.02}  \\ 
\hline
SimCSE-BERT$_{large}$~$\dagger$       & 70.88          & 84.16          & 76.43          & 84.5           & 79.76          & 79.26          & 73.88          & 78.41           \\
DCPCSE-BERT$_{large}$~$\ddagger$      & 73.34          & 85.90          & 77.10          & 85.26          & 80.08          & 80.96          & \textbf{73.28} & 79.42           \\
D2CSE-BERT$_{large}$        & \textbf{74.56} & \textbf{86.53} & \textbf{77.61} & \textbf{85.88} & \textbf{80.64} & \textbf{82.61} & 72.11          & \textbf{79.99}  \\ 
\hline
SimCSE-RoBERTa$_{base}$~$\dagger$     & 70.16          & 81.77          & 73.24          & 81.36          & 80.65          & 80.22          & 68.56          & 76.57           \\
DCPCSE-RoBERTa$_{base}$~$\ddagger$& \textbf{70.57} & 81.91          & 74.60          & 82.9           & 80.96          & 82.84          & 71.70          & 77.93           \\
DiffCSE-RoBERTa$_{base}$~$\ddagger$   & 70.05          & \textbf{83.43} & 75.49          & 82.81          & 82.12          & 82.38          & 71.19          & 78.21           \\
D2CSE-RoBERTa$_{base}$      & 70.53          & 83.22          & \textbf{75.61} & \textbf{83.64} & \textbf{82.53} & \textbf{83.08} & \textbf{72.54} & \textbf{78.74}  \\
\hline
\end{tabular}
}
\label{tab:main}
\end{table*}

Finally, the overall training objective is as follows. \begin{equation}
\mathcal{L} = \mathcal{L_\mathrm{CL}}+\mathrm{\lambda}\mathcal{L_\mathrm{CRTD}}
\end{equation} The regularization coefficient $\mathrm{\lambda}$ is determined empirically.
\section{Experiments}
\label{sec:exp}
\subsection{Setup}
\paragraph{\textbf{Datasets}} For fair comparison, we use the same training datasets as our baselines, which comprise 1M randomly sampled sentences from Wikipedia and NLI including the SNLI~\cite{bowman-etal-2015-large} and MNLI~\cite{williams-etal-2018-broad} datasets for unsupervised and supervised settings, respectively. These datasets are available at the SimCSE's github.\footnotemark\footnotetext{https://github.com/princeton-nlp/SimCSE} As a metric of evaluation, we report the Spearman's rank correlation for the seven STS datasets~\cite{agirre-etal-2016-semeval,cer-etal-2017-semeval,marelli-etal-2014-sick} from SentEval\footnotemark\footnotetext{https://github.com/facebookresearch/SentEval}.

\paragraph{\textbf{Training details.}} We choose our baseline models that have motivated our approach the most, i.e., SimCSE, DCPCSE and DiffCSE with other related methods. 
By carefully following their experimental procedures, we adopt pre-trained checkpoints of \texttt{BERT-uncased-base} and \texttt{RoBERTa-cased} as our frozen sentence encoder and discriminator. More details on our training methodology including hyperparameter optimization and pooling layer choices are presented in the appendix. We note that prompt encoder weights are initialized randomly whereas \texttt{[CLS]} prompt vector is initialized with PLM's embedding weights for the \texttt{[CLS]} token.

\subsection{Results and Analysis}
\label{sec:analy}
\paragraph{\textbf{Semantic textual similarity (STS).}} In Table~\ref{tab:main}, we report the STS performances. Our D2CSE surpasses all baseline models, achieving the highest average Spearman's rank correlation score. 
D2CSE-BERT$_{base}$ only has only about 1\% of the DiffCSE-BERT$_{base}$ parameter size (2.6M vs 220M).
DCPCSE and PromptBERT also are based on prompt learning and performs comparable with D2CSE. 
So, we include the analysis on each model's embedding space in following section.
In supervised setting, we cannot observe nontrivial differences in terms of the evaluation metric between among all methods. We supplement our experimental results on supervised method in the appendix. 

\paragraph{\textbf{Sentence Retrieval.}} We also evaluate on the retrieval task using the STS-B test set and present the result in Table~ \ref{tab:similarity}, \ref{tab:retrievalexample}, \ref{tab:recall}. The most salient part of DiffCSE is its ability to discern subtle differences in meaning. This can be done by training with CRTD tasks. We reason that D2CSE should inherit the same characteristic despite having a fixed discriminator. To prove our point, we have completed the same qualitative analysis on the D2CSE sentence embeddings as done in their paper. 

First, Table~\ref{tab:similarity} shows a query example ``you can do it, too.'' and the corresponding answer with similar candidates. D2CSE results in the highest cosine similarity for a correct answer to the query~\footnotemark\footnotetext{We borrow this example from DiffCSE paper~\cite{diffcse}.} while DCPCSE fails even though both methods equally train on continuous prompts. It is noteworthy that the difference between cosine similarities derived by supervised D2CSE are highly distinctive, whereas supervised DCPCSE and DiffCSE assign high scores on any sentences. Second, Table~\ref{tab:retrievalexample} shows only one example that all methods fail to retrieve the correct answer as the top 1 prediction. And interestingly, two methods trained on CRTD (i.e., D2 and DiffCSE) show exactly the same retrieval results.
Lastly, we report the quantitative performance of retrieval task. We use 97 sentence pairs annotated by 5 out of 5 similarity scores as the query and ground truth answer pairs for retrieval task. 
Table ~\ref{tab:recall} shows the retrieval task results.
As a result, D2CSE shows competitive performance on every metrics, especially on recall@1 metric.

\begin{table}[!ht]
\caption{Cosine similarity scores assigned by each model and method on the given query and candidates. Underline means the max score by each model among the candidates. A check mark refers the correct answer.}
\label{tab:similarity}
    \centering
    \begin{tabular}{lcccc}
    \hline
        Query & \multicolumn{3}{c}{``you can do it, too.''} & Avg. \\
        Candidates & ``you can use it, too.'' & \color{blue}{``yes, you can do it.''(\checkmark)} &  ``can you do it?'' & STS \\ \hline
        \multicolumn{5}{c}{Unpervised Method} \\ \hline
        DCPCSE-BERT$_{base}$ & 0.9017 & 0.8948 & \underline{0.9274} & 78.49 \\ 
        DiffCSE-BERT$_{base}$ & 0.9098 & \underline{0.9233} & 0.9099 & 78.49 \\ 
        D2CSE-BERT$_{base}$ & 0.9343 & \underline{0.9567} & 0.9556 & 79.02 \\ \hline
        \multicolumn{5}{c}{Supervised Method} \\ \hline
        DCPCSE-BERT$_{base}$ & \underline{0.9750} & 0.9729 & 0.9576 & 81.28 \\ 
        DiffCSE-BERT$_{base}$ & 0.9622 & \underline{0.974} & 0.93 & 81.44 \\
        D2CSE-BERT$_{base}$ & 0.9118 & \underline{0.9475} & 0.8663 & 81.52 \\ 
        \hline
    \end{tabular}
\end{table}

\begin{table*}[!ht]
\caption{A retrieval example showing top-k predictions by each model (unsupervised-BERT$_{base}$) from STS-B test set. A check mark refers the correct answer.}
\label{tab:retrievalexample}
\centering
\begin{tabular}{lccc} 
\hline
Query  & \multicolumn{3}{c}{``yes you got it.'' } \\
Answer & \multicolumn{3}{c}{``you got it right.'' }                    \\ 
\hline
Method       & \multicolumn{1}{c}{DCPCSE}            & \multicolumn{1}{c}{DiffCSE} & \multicolumn{1}{c}{D2CSE}  \\ 
\hline
Top1   & yes, you can do it .                            & yes, you can do it .                  & yes, you can do it .                 \\
Top2   & \makecell[c]{yes, you can do exactly\\ what you want to do. } & \color{blue}{you got it right.} (\checkmark) & \color{blue}{you got it right. (\checkmark) }              \\
Top3   & \makecell[c]{yes, that is exactly \\ what it means.}            &              you've got it right.  & you've got it right. \\
Top4   & yes, you should~mention it.  &  \makecell[c]{yes, you can do exactly\\ what you want to do.}   &  \makecell[c]{yes, you can do exactly\\ what you want to do. } \\
Top5   & \color{blue}{you got it right. (\checkmark) }   & \makecell[c]{yes, that is exactly \\ what it means.}  &   \makecell[c]{yes, that is exactly \\ what it means.}                         \\
\hline
\end{tabular}
\end{table*}

\begin{table}
\caption{Retrieval evaluation performance measured on STS-B test set for 3 different models. We build the baseline models by ourselves or download from their GitHub. $\dagger$: built by ourselves, $\ddagger$ : downloaded.}
\label{tab:recall}
\centering
\begin{tabular}{lcccc} 
\hline
\multicolumn{1}{l}{\multirow{2}{*}{Model}} & \multicolumn{3}{c}{Recall} & Avg.   \\
\multicolumn{1}{c}{}                       & @1    & @3    & @5         & STS-B  \\ 
\hline
\multicolumn{5}{c}{Unsupervised Method}                                          \\ 
\hline
DCPCSE-BERT$_{base}$~$\dagger$                       & 90.21 & 96.91 & 98.97      & 78.49  \\
DiffCSE-BERT$_{base}$~$\ddagger$                        & 89.69 & 97.42 & 98.97      & 78.49  \\
D2CSE-BERT$_{base}$                           & 91.24 & 96.91 & 98.97      & 79.02  \\ 
\hline
\multicolumn{5}{c}{Supervised Method}                                            \\ 
\hline
DCPCSE-BERT$_{base}$~$\dagger$                      & 92.78 & 97.94 & 98.45      & 81.28  \\
DiffCSE-BERT$_{base}$~$\dagger$                        & 91.24 & 97.42 & 98.45      & 81.41  \\
D2CSE-BERT$_{base}$                       & 92.78 & 98.45 & 99.48      & 81.52  \\
\hline
\end{tabular}
\end{table}

\begin{figure*}[!t]
\centering
\includegraphics[width=1.0\textwidth]{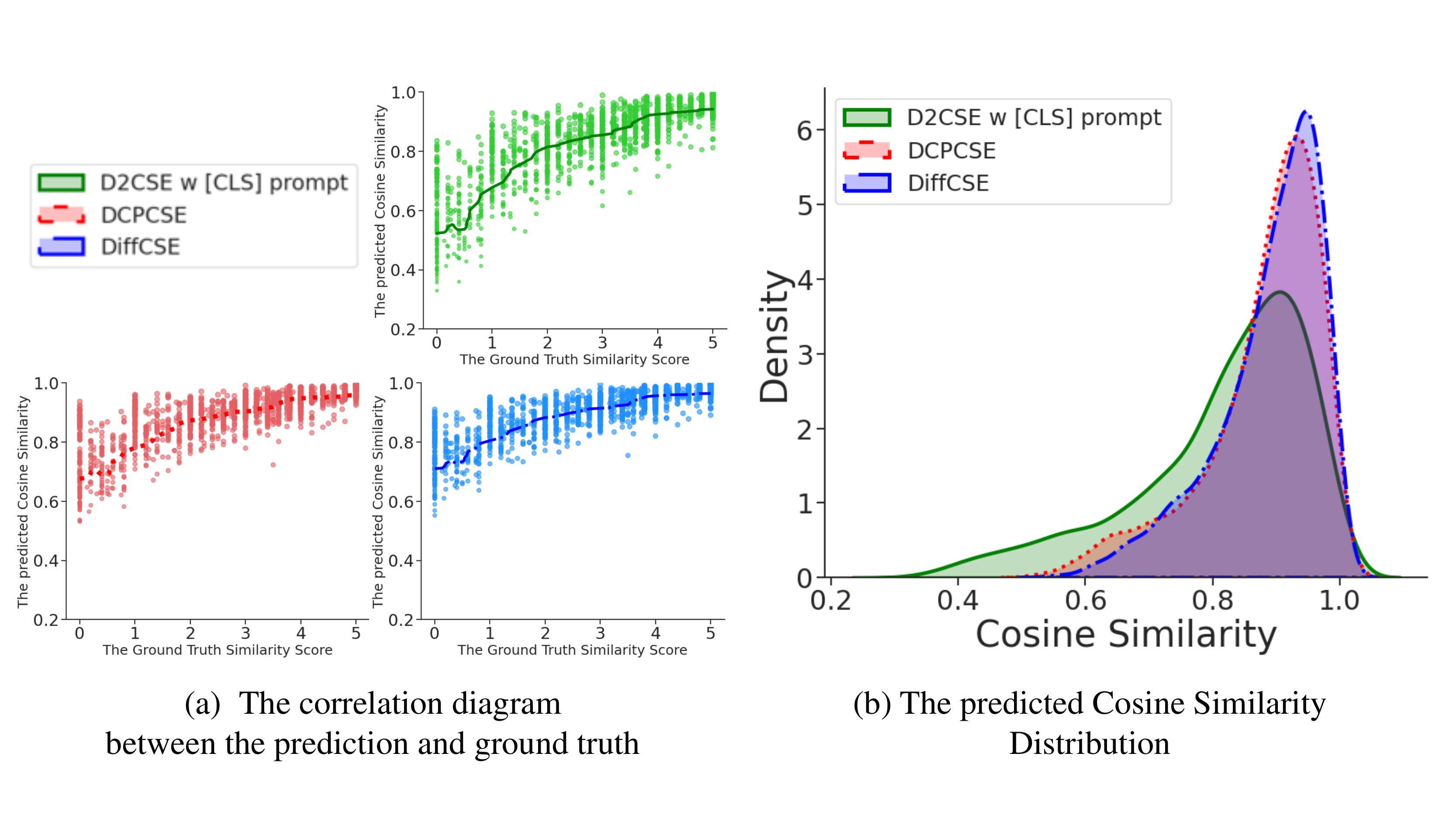}
\caption{Visualization of sentence embeddings for each method. Plot (b) shows the kernel density estimation of the predicted cosine similarity.}
\label{fig:graph}
\end{figure*}

\paragraph{\textbf{Qualitative analysis on sentence embeddings.}} For clarification, we measure $\emph{alignment}$-$\emph{uniformity}$~\cite{understanding} of each model using the STS-B test set and present them in Table~\ref{tab:align-uniform}. The alignment and uniformity metrics are considered a mathematical surrogate to evaluate the quality of sentence representation~\cite{zhang2022mcse}. As the authors of DiffCSE have claimed, we observe the same phenomenon in our unsupervised setting that CRTD brings better alignment while degrading uniformity. Interestingly, however, our supervised method using $\texttt{[CLS]}$ prompt shows notably better $\emph{uniformity}$ while still retaining reasonable $\emph{alignment}$, which means that an anisotropic problem is somewhat alleviated. 
A visualization of this phenomenon can be seen in Figure~\ref{fig:graph}. 
As can be seen from the Figure~\ref{fig:graph}a, b, the sentence embeddings produced by DCPCSE and DiffCSE are distributed in much narrower space than our method.
In our experiment, we have applied the $\texttt{[CLS]}$ prompt to DCPCSE and trained it with a supervised CL objective although we could not have observed such improvements. As the underlying phenomenon requires more in-depth investigation, we leave it for our future work.

\begin{table}[!ht]
\caption{Comparison of alignment and uniformity measured on STS-B test set. For both alignment and uniformity, lower number means better. ($\dagger$ from \cite{diffcse}; $\ddagger$ measured by us; STS averages over all 7 test sets.)}
\label{tab:align-uniform}
\centering
\begin{tabular}{lcc|c}
\hline
\multicolumn{1}{c}{Model} & Alignment & \multicolumn{1}{c}{Uniformity} & \multicolumn{1}{c}{STS} \\ 
\hline
\multicolumn{4}{c}{Unsupervised Method}                                                                         \\ 
\hline
SimCSE-BERT$_{base}$~$\dagger$         & 0.1770    & \textbf{-2.3130}                        & 76.16          \\
DCPCSE-BERT$_{base}$~$\ddagger$        & 0.0842    & -1.5074                        & 78.49                   \\
DiffCSE-BERT$_{base}$~$\ddagger$      & 0.0838    & -1.4593                        & 78.49                     \\
D2CSE-BERT$_{base}$     & \textbf{0.062}    & -1.2627                        & \textbf{79.02}   \\
\hline
\multicolumn{4}{c}{Supervised Method}                                                                        \\ 
\hline
SimCSE-BERT$_{base}$~$\ddagger$         & 0.0898    & -1.7635                       & 81.37          \\
DCPCSE-BERT$_{base}$~$\ddagger$      & \textbf{0.0451}         & -1.4019                              & 81.28                         \\
DiffCSE-BERT$_{base}$~$\ddagger$       & 0.0470         & -1.2857                              & 81.44                         \\
D2CSE-BERT$_{base}$          & 0.0552         & -1.3813                              & 81.29                        \\
+~[CLS] Prompt          & 0.0862          & \textbf{-2.1093}                            & \textbf{81.52}    \\
\hline
\end{tabular}
\end{table}

\begin{figure}[!ht]
\centering
\includegraphics[width=.47\textwidth]{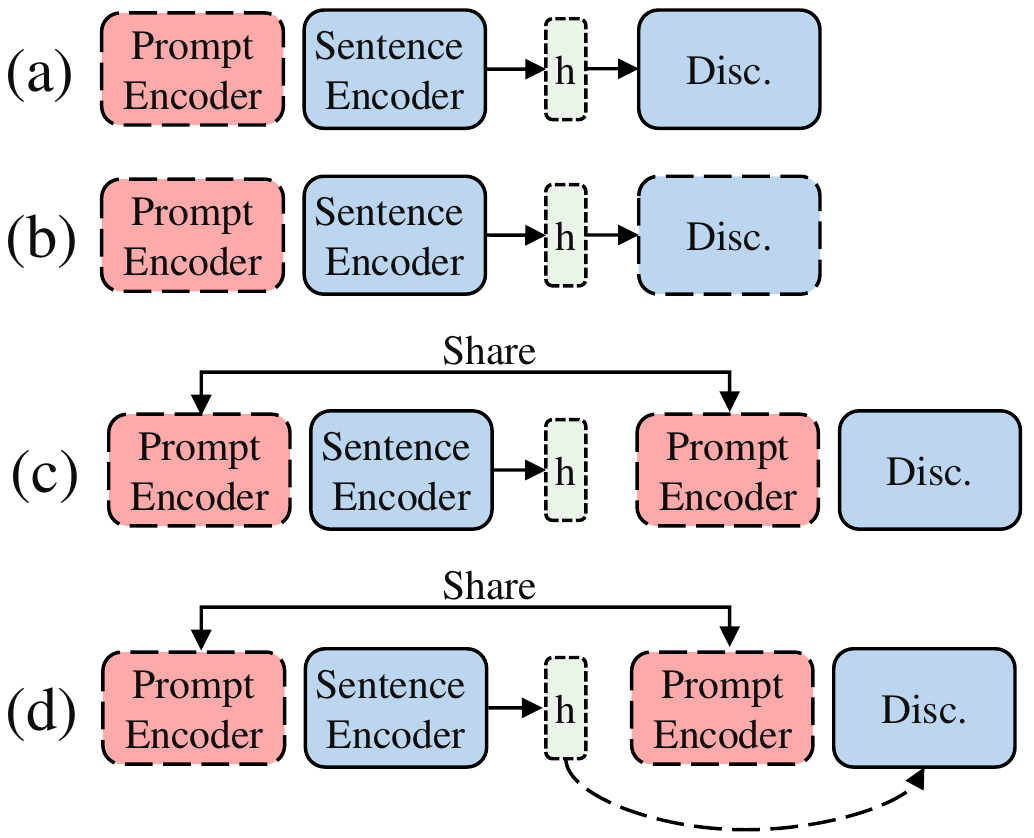}
\caption{Ablations of various frameworks design. `Disc.' refers to the discriminator. Components with dashed lines is trainable, otherwise fixed.}
\label{fig:ablation}
\end{figure}

\subsection{Ablation Study}
\label{sec:ablationstudy}
Practically, a continuous prompt encoder can be applied to either sentence encoder or discriminator, or both.
Due to D2CSE's flexible framework, we can consider a variety of architectures and training methods as shown in Figure~\ref{fig:ablation}.
We describe the details as follows:
\begin{itemize}
  \item (a) applying continuous prompts to only sentence encoder, using the sentence representation $\mathbf{h}$ as a conditional input for CRTD, keeping all PLMs frozen.
  \item (b) applying continuous prompt to only sentence encoder, using the sentence representation $\mathbf{h}$ as a conditional input for CRTD, training a discriminator.
  \item (c) applying shared continuous prompt to both sentence encoder and discriminator, no conditional input, keeping all PLMs frozen.
  \item (d) applying shared continuous prompt to both sentence encoder and discriminator, using the sentence representation $\mathbf{h}$ as a conditional input for CRTD, keeping all PLMs frozen.
\end{itemize}

\begin{table}[!ht]
    \caption{Ablation results of framework design in D2CSE. We report average score on seven STS tasks using D2CSE-BERT$_{base}$ model~(Spearman’s correlation × 100). Each letter (a) through (e) matches the letter in Figure \ref{fig:ablation}. The two columns denote whether we replace static $\texttt{[CLS]}$ token embedding with continuous prompt vector~(dubbed $\texttt{[CLS]}$~prompt).
    }
    \centering
    \begin{tabular}{lcc}
    \hline
        Methods & $\texttt{[CLS]}$prompt & Static$\texttt{[CLS]}$ \\ \hline
        \multicolumn{3}{c}{Unsupervised Method} \\ \hline
        (a) Conditional input only & 78.22 & 78.22 \\ 
        (b) Train discriminator & 78.02 & 78.38 \\ 
        (c) Continuous prompts w/o conditional input & 78.74 & 78.56 \\ 
        (d) Continuous prompts w conditional input & 78.83 & \textbf{79.02} \\ \hline
        \multicolumn{3}{c}{Supervised Method} \\ \hline
        (a) Conditional input only & 81.23 & 81.21 \\ 
        (b) Train discriminator & 81.32 & 81.18 \\ 
        (c) Continuous prompts w/o conditional input & 81.36 & 81.23 \\ 
        (d) Continuous prompts w conditional input & \textbf{81.52} & 81.25 \\
        \hline
    \end{tabular}
\label{tab:ablation}
\end{table}

\paragraph{Results.}
Table~\ref{tab:ablation} shows the results of ablation study. 
From the results of types (a) and (b), the impact of training a discriminator is trivial or even causes a performance drop.
We hypothesize that using both a frozen and trainable PLM together requires more sophisticated hyperparameters and training strategy to boost performance.
In type (c), a discriminator with continuous prompts performs replace token detection task without the conditional input.
In this way, the model is optimized by multiple tasks individually, i.e., contrastive learning and replaced token detection task.
Yet, this architectural choice still outperforms the baseline (78.74 vs 78.48).
Overall, applying prompt encoder to both sentence encoder and discriminator with a conditional input mostly brings performance improvement. 
The two columns denote whether we use `$\texttt{[CLS]}$ prompt'.
As discussed in section~\ref{sec:analy}, `$\texttt{[CLS]}$ prompt' alleviates the anisotropic problem, yet it does not achieve better performance in unsupervised method in terms of STS scores.

\section{Conclusion and Future Work}
We have proposed D2CSE for sentence embeddings by continuous prompts to complement equivariant contrastive learning.
We leverage a newly emerged paradigm, prompt learning, to avoid cumbersome fine-tuning of multiple PLMs done in the predecessor~\cite{diffcse} while substantially improving the quality of sentence embeddings.
Our empirical evaluation strongly indicates that D2CSE outperforms previous work in terms of both STS scores and sentence retrieval task, and also alleviates the anisotropy phenomenon of the embedding space.
Through the qualitative and quantitative analysis on sentence retrieval results, we show that D2CSE can retrieve relevant sentences by capturing subtle semantic similarity, not just based on textual information.
For our future work, we will devise a method that can leverage human-labeled datasets more effectively to boost supervised learning performance. We will also extend D2CSE to various other transfer NLP tasks beyond sentence similarity.


\section*{Appendix}
\label{sec:details1}
\paragraph{Hyperparameters.}
Our framework uses a single Nvidia A100 gpu.
And we implement our code based on SimCSE's code~\footnotemark\footnotetext{https://github.com/princeton-nlp/SimCSE} in Pytorch 1.11.0+cu113.
Importantly, we find that continuous prompt learning method which has limited number of parameter size is highly sensitive to hyperparamters.  
Specifically, the learning rate used in D2CSE and DCPCSE is 4 orders of magnitude larger than DiffCSE, and so it requires more vast searching for the best hyperparameters.
Therefore, we find the best learning rate in two steps. 
First, we search the best value out of [0.01, 0.09], increasing by 0.01.
Second, we set the search range as from the best value to second best one observed in the first step, and search the best value again increasing by 0.001.
And we search the rest of hyperparameters as following ranges:  [128, 256] increasing 16 for batch size, \{0.001, 0.005, 0.01, 0.05, 0.075\} for $\lambda$, \{1, 2, 3\} for epochs under unsupervised setting, \{10, 12, 14, 16, 18, 20\} for prompt length.
Details of all hyperparameters are listed in Table~\ref{tab:hyperun}.

\begin{table}[!ht]
\caption{The best hyperparameters in unsupervised and supervised setting.}
\label{tab:hyperun}
    \centering
    \begin{tabular}{l|ccc|cc}
    \hline
        \multirow{ 2}{*}{Hyperparam} & \multicolumn{3}{c|}{Unsupervised} & \multicolumn{2}{c}{Supervised} \\
         & BERT$_{base}$ & BERT$_{large}$ & RoBERTa$_{base}$ & BERT$_{base}$ & RoBERTa$_{base}$\\ \hline
        Batch Size & 144 & 144 & 128 & 256 & 256\\ 
        Learning Rate & 0.021 & 0.034 & 0.044 & 0.055 & 0.024\\
        Masking Ratio & 0.3 & 0.3 & 0.3 & 0.3 & 0.3\\
        lambda & 0.005 & 0.005 & 0.005 & 0.005 & 0.005\\ 
        Epoch & 2 & 2 & 2 & 10 & 10\\ 
        Prompt Length & 16 & 16 & 16 & 12 & 12 \\ 
        $\mathrm{[CLS]}$ Prompt & False & False & True & True & True\\ \hline
    \end{tabular}
\end{table}

\paragraph{Pooler Design.} 
Following DiffCSE~\cite{diffcse}, we use the two-layer MLP with Batch Normalization~\cite{ioffe2015batch} as our pooler layer.
During training phase, we take the pooling layer's $\texttt{[CLS]}$ token output as the sentence embeddings.
During evaluation phase, however, we take the $\texttt{[CLS]}$ token output before pooling as the sentence embeddings.

\paragraph{Supervised Method Results}
\label{sec:super}
We provide the supervised method results of STS tasks in Table~\ref{tab:supeval}.
Compared to DCPCSE which is only trained by contrastive learning equipped with continuous prompt, D2CSE shows slightly lower performance.
However, we failed to reproduce the results of \cite{dcpcse}.
We build the supervised version of DiffCSE using the same objectives by ourselves because \citet{diffcse} does not consider supervise settings in their work. 
Although D2CSE outperforms DiffCSE, both methods integrating equivariant contrastive learning tend to achieve lower performance than original contrastive learning.


\begin{table*}
\caption{The performance comparison on STS test dataset in supervised setting (Spearman’s correlation × 100). We build the some baseline models by ourselves if they do not provide them. $\dagger$: built by ourselves. $\ddagger$: borrowed from their paper.}
\label{tab:supeval}
\centering
\begin{tabular}{lcccccccc} 
\hline
Model                  & STS12          & STS13          & STS14          & STS15          & STS16          & STS-B          & SICK-R         & Avg.            \\ 
\hline
SimCSE-BERT$_{base}$~$\ddagger$      & 75.30          & \textbf{84.67} & \textbf{80.19} & 85.40          & 80.82          & \textbf{84.25} & 80.39          & 81.57           \\
DCPCSE-BERT$_{base}$~$\dagger$~(reproduce)     & 74.45          & 84.08          & 79.06          & 85.96          & 81.12          & 83.83 & 80.47 & 81.28  \\
DCPCSE-BERT$_{base}$~$\ddagger$      & 75.58          & 84.33          & 79.67          & 85.79          & 81.24          & \textbf{84.25} & \textbf{80.79} & \textbf{81.65}  \\
DiffCSE-BERT$_{base}$~$\dagger$    & \textbf{76.79} & 82.41          & 79.18         & 85.77          & \textbf{82.09} & 83.91          & 79.74          & 81.41           \\
D2CSE-BERT$_{base}$       & 75.08          & 84.34          & 79.80          & \textbf{86.35} & 81.75          & 83.93          & 79.38          & 81.52           \\ 
\hline
SimCSE-RoBERTa$_{base}$~$\ddagger$  & 76.53          & 85.21          & 80.95          & 86.03          & 82.57          & 85.83          & \textbf{80.50} & 82.52           \\
DCPCSE-RoBERTa$_{base}$~$\dagger$~(reproduce)   & 74.71 & 85.96          & 80.42          & 86.13 & 83.19         & 85.76 & 79.70         & 82.27  \\
DCPCSE-RoBERTa$_{base}$~$\ddagger$   & \textbf{76.75} & 85.86          & 80.98          & \textbf{86.51} & 83.51          & \textbf{86.58} & 80.41          & \textbf{82.94}  \\
DiffCSE-RoBERTa$_{base}$~$\dagger$ & 76.37          & 85.59          & 80.65          & 85.34          & 83.44          & 85.79          & 77.99          & 82.17           \\
D2CSE-RoBERTa$_{base}$    & 75.65          & \textbf{87.26} & \textbf{81.39} & 86.31          & \textbf{84.46} & 86.17          & 78.68          & 82.85           \\
\hline
\end{tabular}
\end{table*}

\bibliographystyle{spbasic}
\bibliography{paper}

%
%
%
%




\end{document}